\documentclass[final]{cvpr}

\usepackage[svgnames,table]{xcolor}
\usepackage{times}
\usepackage{epsfig}
\usepackage{graphicx}
\usepackage{color}
\usepackage[pagebackref=true,breaklinks=true,letterpaper=true,colorlinks,bookmarks=false]{hyperref}
\usepackage{amsmath,amsfonts,amssymb,mathrsfs}
\usepackage{makecell}

\usepackage{graphicx}
\usepackage[ruled]{algorithm2e}
\usepackage{amsmath,amssymb}
\usepackage{epstopdf}
\usepackage{tabularx}
\usepackage{eqnarray}
\usepackage{array,booktabs}
\usepackage{enumitem}
\usepackage[space]{grffile} 
\usepackage{soul} 
\usepackage{float}
\usepackage{tabularx,ragged2e} 
\usepackage{bbm}
\usepackage{pifont} 
\usepackage{cancel}
\usepackage{wrapfig}
\usepackage{color, colortbl}

\makeatletter

\renewcommand{\paragraph}{%
  \@startsection{paragraph}{4}%
  {\z@}{0.3ex \@plus 1ex \@minus .2ex}{-1em}
  {\normalfont\normalsize\bfseries}%
}
\makeatother

\definecolor{myGray}{rgb}{0.6,0.6,0.6}
\definecolor{myBlue}{rgb}{0.1,0.1,0.8}

\DeclareMathOperator*{\Var}{Var}

\DeclareMathOperator*{\argmin}{arg\,min}

\def\onedot{\ifx\@let@token.\else.\null\fi\xspace}

\def\eg{\emph{e.g}\onedot} 
\def\ie{\emph{i.e}\onedot}

\def\etal{\emph{et al}\onedot}

\definecolor{pcGray}{rgb}{0.5,0.5,0.5}

\definecolor{pccGray}{rgb}{0.3,0.3,0.3}

\newcommand{\fig}{Fig.~}
\newcommand{\eq}{Eq.\,}
\newcommand{\sect}{Section~}
\newcommand{\tab}{Table~}

\newcommand\minput[1]{%
  \input{#1}%
  \ifhmode\ifnum\lastnodetype=11 \unskip\fi\fi}

\usepackage{listings}
\usepackage{color}

\ifx \@etal \@empty \newcommand{\etal}{et al.} \fi
\ifx \@eg \@empty \newcommand{\eg}{e.g.,~} \fi
\ifx \@ie \@empty \newcommand{\ie}{i.e.,~} \fi
\ifx \@etc \@empty  \fi

\newcommand{\bh}{{\boldsymbol{h}}}

\newcommand{\bw}{{\boldsymbol{w}}}
\newcommand{\bx}{{\boldsymbol{x}}}
\newcommand{\by}{{\boldsymbol{y}}}

\newcommand{\btheta}{{\boldsymbol{\theta}}}

\def\cvprPaperID{11444}
\def\confYear{CVPR 2021}

\newcommand{\tablefont}{\small}
\newcommand{\pp}[1]{\tiny{$\pm${#1}}}
 
\renewcommand{\bw}{{\mathbf{W}}}
\newcommand{\bws}{{\mathcal{W}}} 

\definecolor{myBlue}{rgb}{0, 0, 0.6}
\definecolor{myGreen}{rgb}{0, 0.6, 0}

\makeatletter
\let\@fnsymbol\@arabic
\makeatother

\begin{document}

\title{Unshuffling Data for Improved Generalization}


\author{Damien Teney ~~~~~ Ehsan Abbasnejad ~~~~~ Anton van den Hengel\\
Australian Institute for Machine Learning\\
The University of Adelaide\\
Adelaide, Australia\\
{\tt\small \{damien.teney,ehsan.abbasnejad,anton.vandenhengel\}@adelaide.edu.au}
}

\maketitle
\thispagestyle{empty} 

\begin{abstract}
Generalization beyond the training distribution is a core challenge in machine learning. The common practice of mixing and shuffling examples when training neural networks may not be optimal in this regard. We show that partitioning the data into well-chosen, non-i.i.d. subsets treated as multiple training environments can guide the learning of models with better out-of-distribution generalization. We describe a training procedure to capture the patterns that are stable across environments while discarding spurious ones. The method makes a step beyond correlation-based learning: the choice of the partitioning allows injecting information about the task that cannot be otherwise recovered from the joint distribution of the training data.\vspace{2pt}

We demonstrate multiple use cases with the task of visual question answering, which is notorious for dataset biases. We obtain significant improvements on VQA-CP, using environments built from prior knowledge, existing meta data, or unsupervised clustering. We also get improvements on GQA using annotations of ``equivalent questions'', and on multi-dataset training (VQA v2~/~Visual Genome) by treating them as distinct environments.\vspace{-10pt}
\end{abstract}

\section{Introduction}
\label{sec:intro}

\begin{figure}[t]
  \centering
  \includegraphics[width=1\linewidth]{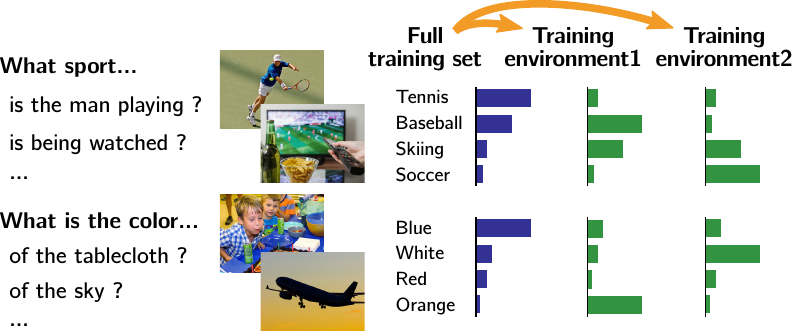}
  \label{fig:teaser}
  \caption{Datasets for visual question answering contain biases and spurious correlations: the first few words of a question are associated with a peaky distribution over answers (\textcolor{myBlue}{blue histograms}).
Models that guess their answers using these correlations generalize poorly. We improve by \textbf{partitioning the data} into multiple training environments across which the spurious correlations vary (\textcolor{myGreen}{green histograms}) while reliable correlations are stable. Our training procedure produces a model that relies on these stable correlations such that it generalizes much better at test time.}
\end{figure}

The best of machine learning models can sometimes be right for the wrong reasons~\cite{agrawal2018don,feng2019misleading,goyal2017making,torralba2011unbiased}. The ubiquitous paradigm of empirical risk minimization (ERM) produces models that capture all statistical patterns present in the training data\footnote{~We use \emph{patterns} and \emph{correlations} interchangeably to refer to statistical relationships between observed random variables, typically the input(s) and output(s) of a supervised learning task.}. However, not all of these patterns are reliable and reflective of the task of interest. Some of them result from confounding factors, sampling biases, and other annotation artifacts specific to a given dataset. We call these patterns \emph{spurious}\footnote{~The literature also uses \emph{dataset biases} to refer to spurious correlations between inputs and outputs that are dataset-specific~\cite{torralba2011unbiased}.} and a model that relies on them will generalize poorly to test data obtained in different conditions (\ie out-of-distribution or OOD data).

The \textbf{limits of ERM} on OOD data are oftentimes overlooked and eclipsed by the common practice of evaluating on test data i.i.d. to the training data~--~a central assumption of classical learning theory~\cite{vapnik1998statistical}. The awareness of these limits has grown with that of their practical implications, from poor transfer across datasets~\cite{torralba2011unbiased} to biases and fairness issues~\cite{adeli2019bias} and vulnerability to adversarial inputs~\cite{grand2019adversarial}. As a result, benchmarks with OOD test sets are becoming increasingly common in vision and NLP~\cite{agrawal2018don,akula2020words,barbu2019objectnet,guo2019quantifying,jia2017adversarial,li2017deeper}. This paper presents a training paradigm to improve OOD generalization.

The study of \textbf{generalization in computer vision} has a long history~\cite{evgeniou2004regularized,khosla2012undoing,mitchell1980need}. The rise in popularity of high-level tasks like visual question answering (VQA)~\cite{antol2015vqa}, visual dialogue~\cite{das2017dialog}, or vision-and-language navigation~\cite{anderson2018vision} has made the topic even more important. The complexity of these tasks and the combinatorial explosion of the size of their input domain make it impossible to process training data densely spanning this space. Models trained with ERM are then more likely to latch on spurious correlations (because they are often easier to fit~\cite{shah2020pitfalls}) rather than on the true reasoning process that underlies the task~\cite{jabri2016revisiting}. VQA was shown empirically to be a prime example of this issue. Many methods have been proposed to address it~\cite{agrawal2018don,cadene2019rubi,clark2019don,grand2019adversarial,guo2019quantifying,mahabadi2019simple,ramakrishnan2018overcoming}.

We propose a \textbf{general method to improve OOD generalization}. We discourage the model from using spurious correlations that only appear in subsets of the training data, and rather ensure that it uses reliable ones that are more likely to generalize at test time.
More precisely, we first partition the data into multiple training environments~\cite{arjovsky2019invariant} such that spurious correlations vary across environments while reliable ones remain stable.
We later describe multiple strategies to build such environments, using unsupervised clustering, prior knowledge, and auxiliary annotations in existing datasets.
Second, we train multiple copies of a neural network, one per environment. Some of their weights are shared across environments, while others are subject to a variance regularizer in parameter space.
This leads the model to extract features that are stable across environments (\ie features that do not represent environment-specific properties) since they are optimized to be predictive under a classifier common to all environments (as encouraged by the variance regularizer). Additional intuitions and reasons why this approach is superior to ERM are discussed in \sect\ref{sec:why}.

We provide \textbf{empirical evidence of improvements} in three distinct use cases on the task of VQA. First, we demonstrate improved resilience to language biases with VQA-CP~\cite{agrawal2018don}. Second, on GQA~\cite{hudson2018gqa}, we show how to use annotations of equivalent questions (some training questions being rephrasings of others). We obtain substantial gains over simple data augmentation with these equivalent questions in a small-data regime. Third, we show a small benefit in training a model on multiple datasets by treating VQA v2~\cite{goyal2017making} and Visual Genome QA~\cite{krishnavisualgenome} as two environments rather than one aggregated dataset. Of these three use cases, the first is the most well-studied but \textbf{our method has a much wider scope than the VQA-CP dataset}.

\vspace{3pt}
\noindent
The contributions of this paper are summarized as follows.
\setlist{nolistsep,leftmargin=*}
\begin{enumerate}[noitemsep]
  \item We propose to partition existing datasets into training environments to improve generalization. We describe the requirements for the partitioning and a procedure to train neural networks to rely on stable correlations across environments while ignoring spurious ones.

  \item We apply the method to three use cases with the task of VQA: (1)~resilience to language biases, (2)~leveraging known relations of equivalence between specific training questions, and (3)~multi-dataset training.

  \item We provide empirical evidence of clear improvements and a extensive sensitivity analysis to hyperparameters and implementation choices.
\end{enumerate}

\section{Related work}
\label{sec:related}

This paper touches on fundamental aspects of machine learning and so is related to many existing works. 

\paragraph{Dataset biases in vision-and-language tasks.}
Several popular datasets used in vision-and-language~\cite{goyal2016balanced} and natural language processing~\cite{zellers2018swag} have been shown to exhibit strong biases.
A model trained naively on these datasets can exhibit surprisingly good performance by relying on dataset-specific biases  without capturing the true mechanisms of the task.
On the evaluation side, there is a trend towards out-of-distribution test data to better identify this behaviour (\eg~\cite{agrawal2018don,akula2020words,barbu2019objectnet,guo2019quantifying,jia2017adversarial,li2017deeper,zellers2018swag}).

\paragraph{Invariances and generalization.}
Resilience to dataset biases cannot be solved by simply collecting more data from the same distribution, since it would still contain the same unreliable patterns.
The data collection process can be improved~\cite{goyal2016balanced,zellers2018swag,zhang2015balanced} but this option only addresses precisely identified biases and confounders.
Improving generalization requires to bring in information (often implicitly) about the mechanisms of the task of interest that go beyond what is represented by the joint distribution of the training data~(see discussion \sect\ref{sec:why}). Common methods include architecture design and data augmentation~\cite{krizhevsky2012imagenet,tanner1987calculation} to specify input transformations the model should be invariant to. This helps ignoring spurious correlations and improves generalization but it requires explicit knowledge of desirable invariances.
In comparison, our method discovers invariances implicitly. We train a model to rely on input features that are similarly predictive (\ie invariant) across environments. The required expert knowledge is displaced to the specification of the partitioning into environments.

\paragraph{Aggregating datasets.} Using training data collected in different conditions is often beneficial for generalization (\eg in \cite{teney2017challenge} for VQA) because biases in each dataset are likely different and they can ``cancel each other out''. We argue however that treating aggregated datasets as a collection of samples from a unique distribution loses valuable information. Our method treats them as distinct training environments and seeks to identify patterns that are similarly predictive across them while discarding those that are dataset-specific.
Khosla \etal~\cite{khosla2012undoing} also showed that accounting for bias when combining datasets was beneficial for generalization.

\paragraph{Domain adaptation and domain generalization.} 
Training a model under multiple environments is reminiscent of domain adaptation~\cite{ganin2016domain}. But our objective is not to adapt to one particular target domain but rather to generalizes across a range of unknown conditions. Our setup is more similar to the recently-introduced terminology of domain generalization~\cite{wang2019learning,gulrajani2020search,wang2019learning}. These methods specifically target image recognition benchmarks like PACS and VLCS.


\paragraph{Ensembles.}
Our method trains multiple copies of a model in parallel, which is superficially similar to ensembling~\cite{zhou2012ensemble} and bootstrap aggregation a.k.a. bagging~\cite{Breiman1996}. Traditional ensembles however combine models in output space. We combine models in parameter space\footnote{~Averaging predictions or final weights is equivalent with a linear classifier. It is not in our case of a non-linear output.}.
We show empirically that the improvements from our approach are distinct from (and complementary to) those of traditional ensembling. Bagging uses uniform sampling, whereas the point of our method is to exploit prior knowledge to build the training environments.

\paragraph{Robustness in VQA.}
State-of-the-art models for VQA have been shown to be strongly reliant language biases.
Benchmarks have been designed to better study the issue~\cite{agrawal2018don,hudson2018gqa,johnson2016clevr}. VQA-CP~\cite{agrawal2018don} allows out-of-distribution evaluation, where the joint distribution of questions and answers is different in the training and test sets. Methods have been proposed with strong improvements on VQA-CP~(\cite{cadene2019rubi,clark2019don,grand2019adversarial,guo2019quantifying,mahabadi2019simple,ramakrishnan2018overcoming} among others).
However, many were been shown to cheat the evaluation by exploiting knowledge of the construction of VQA-CP that should have been kept private~\cite{teney2020value}.
In \cite{teney2020learning}, the authors use counterfactual examples to learn which input features to focus on. This is comparable to our method in that we learn by contrasting training environments, whereas \cite{teney2020learning} proceeds at the instance level.
In \cite{teney2019incorporating}, the authors exploited auxiliary annotations of the GQA dataset to improve robustness, which we also use in some of our experiments.

\paragraph{Fair and bias-resilient machine learning.}
Addressing dataset biases is also motivated by issues of fairness~\cite{adeli2019bias,hendricks2018women,wang2018adversarial,zhao2017men}. These works aim to build predictive models that are invariant to specific attributes of the input, such as gender or ethnicity. These attributes need to be specified and annotated, which is very limiting. For example in VQA, there is a known desired invariance to some linguistic patterns in the question, but their exact form is not known and cannot be annotated as a discrete attribute.

\paragraph{Invariant risk minimization.}
This paper is strongly inspired the principle of invariant risk minimization (IRM)~\cite{arjovsky2019invariant}. Arjovsky \etal showed how training under multiple environments can improve OOD generalization. Our contributions over~\cite{arjovsky2019invariant} are threefold: an alternative, easy-to-train implementation, an application to a real large-scale dataset, and demonstrations of how to obtain training environments. Other implementations were proposed~\cite{ahuja2020invariant,chang2020invariant,Choe2020AnES,idnanilearning} but mostly demonstrated on toy data. Theoretical and empirical comparisons of variants of IRM are missing pieces to address in the future. Creager \etal \cite{creager2020environment} attempted unsuccessfully to discover environments automatically for IRM. Their failure, even on toy data, supports the view that the true source of improvements is in the information used to obtain or create the environments.

\paragraph{Learning from groupings of data} has been studied from statistical \cite{bouchacourt2017multi,heinze2017conditional} and causal \cite{pashami2018causal,sgouritsa2013identifying} points of view.
Heinze-Deml and Meinshausen~\cite{heinze2017conditional} used a variance regularizer on predictions across versions of an example, such as multiple photos of a same individual. Our variance regularizer, in comparison, acts on the parameters of the model. And we do not require correspondences between specific training examples.
\cite{evgeniou2004regularized} is another classic work using a variance regularizer for multi-task learning.
And Vasilescu \etal~\cite{vasilescu2002multilinear} were among the first to compute invariant representations for the causal factors of image formation, which they did for face recognition and human motion signatures.

\section{Proposed approach}

\subsection{Partitioning data into training environments}
\label{sec:partitioning}
The main intuition behind our method is that the training data contains both reliable and spurious correlations between inputs and labels, and that it is sometimes possible to partition the data into ``training environments'' in which the strength of the spurious ones is affected more than the reliable ones.
We then train a model to rely on the correlations that are stable across environments. The corollary is that it ignores the environment-specific spurious ones.

As an example in VQA, let's imagine the question \textit{What animal is in the picture ?} A reliable correlation is the presence of canines in images that have \text{dog} as answer. An example of a spurious correlation is that most questions starting with \textit{What sport...} have \textit{tennis} as answer. A model that relies on this correlation irrespective of image contents or of the rest of the question will generalize poorly.
This spurious correlation results from annotator and selection biases. Conceivably, data collected from two groups of annotators will exhibit different biases, \eg one group mostly picking images with tennis as the answer, the other football.
Our approach identifies such groupings of the data as training environments then trains a model that uses the correlations that are stable (\ie similarly predictive) across environments.

Concretely, we partition the training set 
$\mathcal{T}{=}\{(\bx_i,\by_i)\} _i$ 
of inputs $\bx_i$ and labels $\by_i$ (one-hot vectors in a classification task) into $E$ disjoint training environments ${\mathcal{T}_e}$ such that 
$\bigcup_{e=1}^E \mathcal{T}_e=\mathcal{T}$. 
The environments are built to isolate the effect of spurious correlations, such that only the strength of reliable correlations remains stable across all environments.
We provide additional justification for the principle in \sect\ref{sec:why} and we describe strategies to build environments from existing datasets in \sect\ref{sec:experiments}. We show that they can be built by unsupervised clustering, by injecting prior knowledge, and by leveraging auxiliary annotations from existing datasets.
Next, we describe how to train a model across environments to rely on stable correlations while ignoring spurious ones.


\subsection{Training over multiple environments}
\label{sec:training}

Our goal is to learn a predictive model $\Phi$ that maps an input $\bx$ to an output $\hat{\by}\,$=$\,\Phi(\bx)$ such as a vector of class probabilities in a classification task. 
We represent the model as the combination of a feature extractor and a subsequent linear classifier.
The feature extractor $f_\btheta(\bx)$ (typically with a deep neural network) uses parameters $\btheta$ to extract a vector $\bh\,{=}\,f_\btheta(\bx)$.
The subsequent linear classifier is a matrix of weights $\bw$ and the whole model is described as 
$\Phi(\bx)\,{=}\,\bw f_\btheta(\bx)\,$.
The standard training procedure is to optimize $\btheta$ and $\bw$ for maximum likelihood on the training set $\mathcal{T}$ under a loss $\mathcal{L}$, \ie solving the following optimization problem:
\begin{equation}
  \argmin_{\btheta,\bw} \; \Sigma_{(\bx,\by) \in \mathcal{T}} \, \mathcal{L}\big( \bw f_\btheta(\bx), \by\big) ~.
\end{equation}

In our method, assuming a prior definition of training environments $\mathcal{T}_e$ (see \sect\ref{sec:partitioning}), we want to train the model to be highly predictive on the training environments as well as on an OOD test set, in which only the input/output correlations \emph{common to all training environments} can be assumed to hold. We train a different model $\Phi_e(\bx)\,{=}\,\bw_e\,f_\btheta(\bx)$ for each environment. The feature extractor $f_\btheta(\cdot)$ is shared, such that it identifies features common to all environments. But a different classifier $\bw_e$ is optimized for each environment. We want the features extracted by $f_\btheta(\cdot)$ to be stable, \ie similarly predictive across environments. For this, we choose to encourage the parameters of the classifiers $\bw_e$ to converge to a common value. This is naturally implemented by minimizing their variance over $e\,{=}\,1..E$.

At test time, we use $\Phi^\star(\bx)\,{=}\,\bar{\bw} \, f(\bx)$, where $\bar{\bw}$ is the arithmetic mean of $\bw_e$ over $e$. 
Since the variance regularizer brings all $\bw_e$ toward a common value during training, the arithmetic mean is a natural choice.
The complete optimization task is defined as:
\begin{equation}
  \argmin_{\btheta,\bws} \Sigma_e \;\Sigma_{(\bx,\by) \in \mathcal{T}_e} \;\mathcal{L}\big( \bw_e f_\btheta(\bx), \by\big) \;+\; \lambda \Var_e(\bw_e)
  \label{eq:loss}
\end{equation}
where $\lambda$ is a scalar hyperparameter, $\bws = \{\bw_e\}_{e=1}^E$, and $\Var_e(\bw_e)$ is the variance of classifier weights. The standard definition of the variance gives
\begin{align}
  &\Var_e(\bw_e)~ = ~(1/E) \;\Sigma_e ~||\bw_e - \bar{\bw} ||^2\\
  &\textrm{with}~~~ \bar{\bw} = (1/E) \; \Sigma_e \bw_e~.
\end{align}
We refer to this definition as the ``absolute variance'' in our experiments. Finding a unique best value for $\lambda$ in \eq\ref{eq:loss} proved difficult because the magnitude of the weights can vary widely during the early stages of the optimization. As a remedy, we use a relative measure of variance that rescales each term by the weights' magnitude:
\begin{align}
  &\Var_e(\bw_e)~ = ~(1/E) \;\Sigma_e ~\big(||\bw_e - \bar{\bw} ||_2 / \|\bw_e\|_1\big)^2
\end{align}
It slightly improves our results (see \tab\ref{tabVqaCpAblation}) and makes $\lambda$ easier to tune.
We also found small improvements in optimizing \eq\ref{eq:loss} alternatively: one mini-batch serves to update $\btheta$, another one to update $\bw$, repeating until convergence. It slightly improves the final accuracy but it is not crucial to the method and was only used in select experiments reported in \tab\ref{tabVqaCpAblation}.

\begin{figure}[t!]
  \centering
  \includegraphics[width=0.99\linewidth]{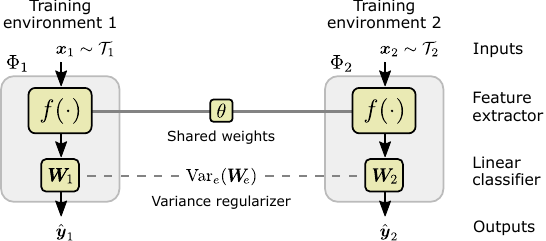}
  \vspace{4pt}
  \caption{During training, we optimize a different copy of the model under each environment (each sees a different subset of the data). Two environments are pictured but our experiments use up to 18. The objective is to have the model rely on statistical patterns that are stable across environments. The weights of the feature extractor ($\btheta$) are shared across environments, those of the classifier ($\bw_e$) are not. A regularizer encourages the latter to converge to a unique solution simultaneously optimal across environments. At test time, we use the arithmetic mean of these weights $\bar{\bw}$.}
  \label{fig:overall}
\end{figure}

\subsection{Why it works}
\label{sec:why}

This section discusses further connections with existing methods and formal interpretations of generalization.

\paragraph{Causal view of generalization.}
Improving generalization requires distinguishing dataset-specific artifacts from the actual mechanisms of the task of interest. A causal view of generalization formalizes this as identifying causal properties\footnote{~We do not claim to identify a causal model, we only aim at producing a predictive model. The causal formalization only serves to highlight that properties relevant to OOD generalization cannot be recovered from the joint distribution represented by a standard dataset, but can be informed by training across well-chosen environments.} of the data-generating process behind the observed data~\cite{arjovsky2019invariant,peters2015causal}. A critical outcome of this formalization is that such causal properties are provably impossible to recover solely from the joint distribution over inputs and outputs that a standard training set represents. In other words, spurious and reliable correlations cannot be distinguished from one another.

Common avenues for bringing the missing information include architecture design (setting a prior on the causal structure), data augmentation (defining invariances in the input domain) and other task-specific inductive biases. Our method falls in another category that seeks to exploit implicit training signals found unexploited in existing datasets~\cite{arjovsky2019invariant,teney2020learning,heinze2017conditional,kaushik2019learning}. More specifically, we use multiple training environments. These emerge naturally in data collected from multiple annotators, sites, or annotation interfaces. Data points are usually collated into a single dataset, but this loses information. Our premise is that such environments elicit different spurious correlations without affecting the reliable patterns inherent to the task of interest.

The fact that observational data from multiple environments can be informative about causal properties of the data generating process may seem at odds with basic principles of causal reasoning~\cite{pearl2000causality}. This is resolved when considering each environment as an intervention on the data generating process.
Assuming these interventions only act on variables spuriously correlated with the output (hence the importance of a well-chosen partitioning), the causal mechanisms between non-intervened variables and the output will remain unchanged by the principle of independent causal mechanisms~\cite{peters2017elements}. The statistical patterns inherent to the task thus remain stable across environments.

\paragraph{Source of improvements.}
The improvements obtained through our method therefore hinge on the information used to obtain training environments (\eg multiple existing datasets) or to create them by ``unshuffling'' an existing dataset. The latter requires the practitioner to bring in meta knowledge about the task or dataset, such as axes along which to cluster the data.
Conversely, environments made as random partitions are not expected to bring any benefit, as verified in our experiments (see \tab\ref{tabVqaCpAblation}).

\paragraph{Invariant risk minimization.}
Our training procedure is inspired by the principle of invariant risk minimization (IRM)~\cite{arjovsky2019invariant}. IRM proposes to identify a representation of data such that the optimal classifier, on top of this representation, is identical across environments. Formally, using our notations, this amounts to optimizing the feature extractor $f_\btheta(\cdot)$ and linear classifier $\bw$ for the following objective:
\begin{align}
  &\min_{\btheta,\bws} ~\Sigma_e \; \Sigma_{(\bx,\by) \in \mathcal{T}_e} \; \mathcal{L}\big(\bw^\star f_\btheta(\bx), \by\big) \\
  &\textrm{s.t.}~\bw^\star \in \argmin_\bw \Sigma_{(\bx,\by) \in \mathcal{T}_e} \; \mathcal{L}\big(\bw f_\btheta(\bx), \by\big), ~\forall\; e.
  \label{eq:irm}
\end{align}
The constraint on $\bw^\star$ is the crux of the principle. A classifier that is optimal in a given environment can only use the features that are reliable predictors in that environment. Requiring the classifier $\bw^\star$ to be simultaneously optimal across all environments (\ie at the intersection of all environment-specific optima) means that it can only use stable features. In other words, consider a spurious correlation, specific to an environment $e$, between the output labels and a feature $\tilde{\bx}$. A model (feature extractor and classifier) trained in isolation on $e$ would use this feature $\tilde{\bx}$. However, this spurious correlation does not hold in another environment $e'$. Even though the shared feature extractor could extract some semblance of the feature $\tilde{\bx}$ in $e'$, this feature will not be predictive in the same way as in $e$. Therefore, the optimal classifier in $e'$ will not use $\tilde{\bx}$ in the same way. Since we are looking for a unique classifier that is simultaneously optimal in $e$ and $e'$, the shared feature extractor must ignore this unreliable feature, and only extract those that are \emph{similarly predictive} across environments.

\paragraph{Proposed method and IRM.}
The objective of \eq\ref{eq:irm} involves an impractical nested optimization. The approximation proposed in \cite{arjovsky2019invariant} replaces the constraint with a regularizing term in the objective that uses the gradient of the environment-specific risk with respect to the classifier. The resulting objective is highly convex and has been reported to be difficult to train in practice. Our method (\eq\ref{eq:loss}) uses the variance of $\bw_e$ as a regularizer. The gradient of the risk in \cite{arjovsky2019invariant} is motivated as a measure of ``how optimal'' a classifier is. Our version operates directly in the parameter space of the classifier.
As explained in \sect\ref{sec:training}, our version intuitively leads to stationary points that satisfy the IRM principle, but further work is warranted to study its convergence properties and possible guarantees discussed in~\cite{arjovsky2019invariant}. Our results are only empirical but showed it to be very stable during training and highly effective in our use cases.

Finally, recent breakthrough theoretical~\cite{khemakhem2020variational} and empirical results~\cite{sorrenson2020disentanglement} proved identifiability in non-linear ICA when the generating process is conditional distribution of which the conditioning variable is observed. This setting is analogous to training across environments with the environment ID interpreted as the conditioning variable. Further work is needed to establish formal connections with these results.

\section{Experiments}
\label{sec:experiments}

We present three applications on the task of VQA, which is notorious for dataset biases.
Our strongest results are with the VQA-CP dataset~\cite{agrawal2018don} which is designed to test out-of-distribution generalization. The other two application use GQA~\cite{hudson2018gqa}, and VQA v2~\cite{goyal2016balanced} combined with Visual Genome QA~\cite{krishnavisualgenome}. The quantitative improvements in these other two applications are smaller but they demonstrate the wider applicability of the method. Most other methods in \tab\ref{tabVqaCpComparison} are specific to VQA-CP.

\paragraph{Implementation.}
We implemented the method on top of the ``bottom-up and top-down attention'' model~\cite{teney2017challenge} (details in supp. mat.) since it serves as the baseline of most competing techniques on VQA-CP~\cite{cadene2019rubi,clark2019don,grand2019adversarial,guo2019quantifying,mahabadi2019simple,ramakrishnan2018overcoming}. Our method should readily apply to recent models~\cite{chen2019uniter,gao2019dynamic,gao2019multi,li2019unicoder,liu2019learning,tan2019lxmert,yu2019deep} including those with stronger baseline performance on GQA~\cite{hu2019language,hudson2018compositional}. Evaluations on VQA-CP follow the guidelines of Teney \etal~\cite{teney2020value}, including performing ablations on ``Other'' questions only, and reporting both in-domain and OOD performance without retraining. Our results are also averaged across multiple runs (thus not to outcome of one possibly-lucky random seed) unlike most results reported in the literature.

\subsection{Robustness to language biases (VQA-CP)}
\label{sec:vqaCp}

\paragraph{Experimental setup.}
The VQA-CP dataset~\cite{agrawal2018don} was constructed by reorganizing VQA v2~\cite{goyal2016balanced} such that the correlation between the question type and correct answer differs in the training and test splits. For example, the most common answer to questions starting with \textit{What sport...} is \textit{tennis} in the training set, but \textit{skiing} in the test set. A model that guesses an answer primarily from the question will perform poorly. In our experiments, we report the accuracy on the official test set, but also on a validation set that we built by holding out 8,000 random instances from the training set. This serves as to measure ``in-distribution'' performance, while the test set serves to measure generalization to out-of-distribution data. As discussed in~\cite{teney2020value}, evaluation on the `yes/no' and `number' categories of VQA-CP have unintuitive issues (for example, randomly guessing yes/no on the former category achieves 72.9$\%$ while a method like~\cite{agrawal2018don} only gets 65.5$\%$; thus, a random, untrained model is usually better than a trained one). For these reasons, our ablation study uses only the `other' type of questions.

\begin{figure}[t]
  \centering
  \includegraphics[height=0.41\linewidth]{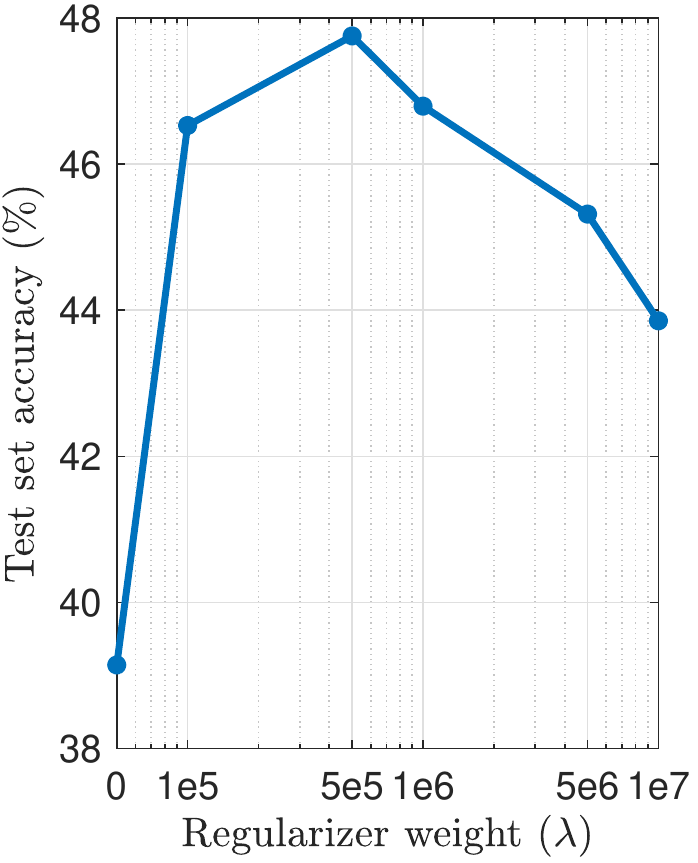}~\includegraphics[height=0.41\linewidth]{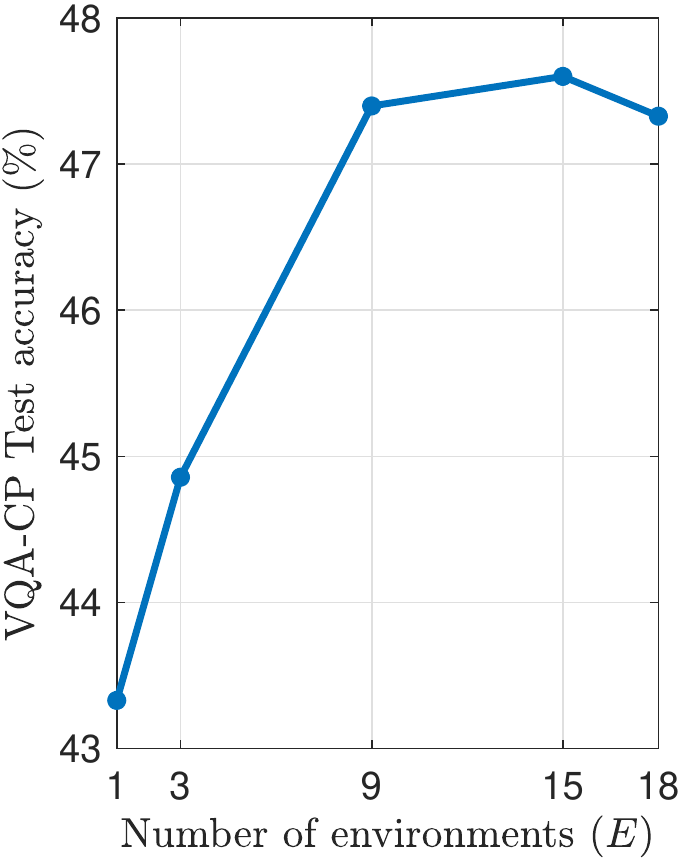}~\includegraphics[height=0.41\linewidth]{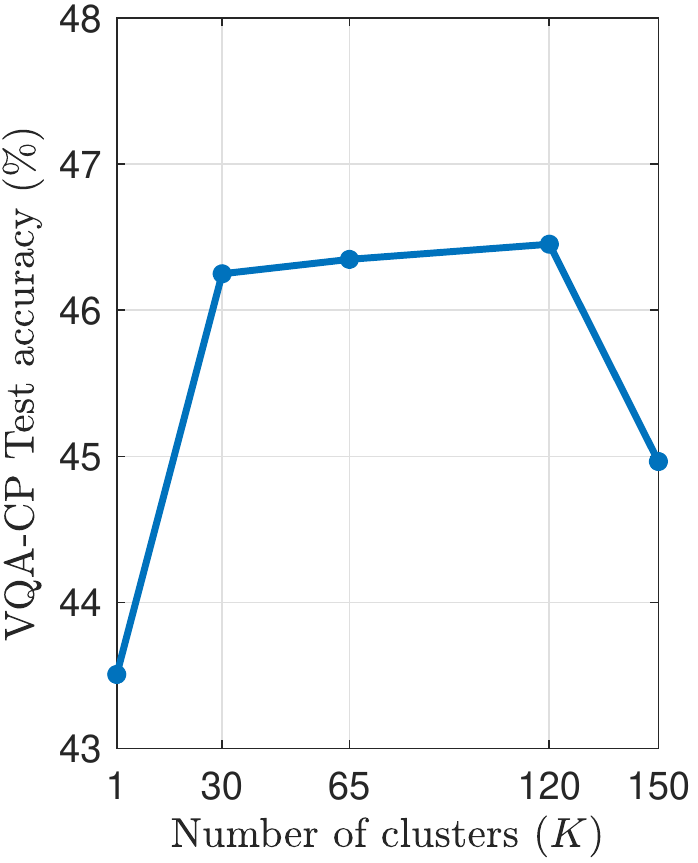}
  \caption{Sensitivity to hyperparameters on VQA-CP, using environments built from question groups (left and middle) or by clustering questions (right). See discussion in \sect\ref{sec:vqaCp}.\label{fig:vqaCp}}
\end{figure}

\vspace{5pt}
\paragraph{Environments from ground truth question types.}
We first present experiments for which we built training environments with the ground truth type of questions (provided with the dataset). Each training question has one label among 65. This label serves as a natural clustering of the data. We assign the 65 clusters randomly to $E$ environments, splitting clusters as needed to obtain the same number of training questions per environment. We trained our method with a different number of environments (see \fig\ref{fig:vqaCp}b). The point $E{=}1$ corresponds a standard training of the model with the whole dataset. The plot shows a clear improvement with multiple environments, with a peak performance with $E$=15. Why does the accuracy decrease with more environments ? We believe that the diversity and amount of data in each environment then gets too low. We experimented with other strategies (not reported in plots and tables) to assign clusters to environments other than randomly, by maximizing or minimizing the variation in the answer distribution in each environment (compared to the whole dataset). We found that the random assignment performed best. It keeps the distribution of answers relatively similar across environments, unless $E$ is too large, which further explains the slight decrease in accuracy then.

\vspace{5pt}
\paragraph{Environments by clustering questions.}
We now present experiments where the environments are built through unsupervised clustering of the questions. We do not use the ground truth question types here. We rely on our prior knowledge that a model should not be overly reliant on the general form of a question. We represent the questions as binary bag-of-words vectors (details in supp. mat.) and cluster them with $K$-means. As above, we then assign the clusters randomly to $E$ environments ($E < K$). We plot in \fig\ref{fig:vqaCp}c the accuracy of the model against the number of clusters $K$. There is a distinct broad optimum. The best accuracy is close but still inferior to the strategy that uses the ground truth question types (compare the peaks in \fig\ref{fig:vqaCp}b and c). We measured the similarity of the unsupervised clustering with the ground truth type in terms of Rand index, and noted that it was positively correlated with the accuracy. This shows that using ground truth types is the better strategy and the clustering approximates it.

\vspace{5pt}
\paragraph{Ablative analysis.}
We provide an ablation study in \tab\ref{tabVqaCpAblation}. The performance substantially increases on the test set with the proposed method compared to all baselines. The variance regularizer is crucial to the success of the method. We plot in \fig\ref{fig:vqaCp}a the accuracy as a function of the regularizer weight ($\lambda$ in \eq\ref{eq:loss}). There is a clear optimum, with higher values being generally better (the plot uses a log scale). In \tab\ref{tabVqaCpAblation}, we also observe that the relative variance performs slightly better than the absolute variance. We also note that the alternating optimization scheme performs slightly better. It works best after a a few epochs of non-alternating ``warm-up'', during which all parameters are updated together. The use of the alternating optimization is not crucial to the overall success of the method, and it is not used in any other experiment.

\vspace{5pt}
\paragraph{Comparison to existing methods.}
We trained our method on the whole VQA-CP dataset, including `yes/no' and `number' questions to compare it against existing methods (see~\tab\ref{tabVqaCpComparison}). Our method surpasses all others on `other', most of them by a large margin. The method of Clark \etal~\cite{clark2019don} gets better results on the `yes/no' and `number' questions, but its results on the standard splits of VQA v2 are also down to baseline levels (\ie similar to a random guess out of the subset of answers used in each category). In comparison, our performance on the standard splits remains higher. Note that some competing methods admittedly use the test set as a validation set (!) for hyperparameter selection and/or model selection~\cite{agrawal2018don,grand2019adversarial}. We rather hold out 8k instances from the training set to serve as a validation set as recommended in~\cite{teney2020value}. They serve for example to monitor training and determine the epoch for early stopping.

\begin{table}[t]
\footnotesize
\renewcommand{\tabcolsep}{0.22em}
\renewcommand{\arraystretch}{1.25}
\centering
\begin{tabularx}{\linewidth}{X cc}
\Xhline{1\arrayrulewidth}
                                          & \multicolumn{2}{c}{VQA-CP v2} \\ \cline{2-3}
                                          & Val. set & Test set \\
                                          & Other & Other \\
\Xhline{1\arrayrulewidth}
  Baseline & 54.74 & 43.33 \\
  Environments: random; \hspace{4.6em}rel. var., no alt. opt. & 53.34 & 43.51\\
  Environments: clustered questions; \hspace{0pt}rel. var., no alt. opt. & 54.10 & 46.35 \\
  \textbf{Environments: question groups}; \hspace{0.38em}rel. var., no alt. opt. & 53.87 & 47.60 \\
  ~~+ Alternating optimization (0 warm-up epoch) & 54.00 & 47.71 \\
  ~~+ Alternating optimization (2 warm-up epochs) & 53.90 & 47.82 \\
  ~~\textbf{+ Alternating optimization (4 warm-up epochs)} & 53.98 & \textbf{48.06} \\
  ~~+ Alternating optimization (6 warm-up epochs) & 53.86 & 47.38 \\
  ~~Without variance regularizer & 40.76 & 39.14 \\
  ~~With absolute variance regularizer  & 51.44 & 46.17 \\
\Xhline{1\arrayrulewidth}
\end{tabularx}
\normalsize
\normalsize
\caption{Ablative study on VQA-CP (accuracy in percent, training on `Other' questions only). Our method brings a significant gain over the baseline, both with environments built using the ground truth question types, and with environments built by unsupervised clustering of the questions. As a sanity check, we run the method with random environments, which gives results essentially identical to the baseline, as expected. The alternating optimization scheme brings a additional small improvement, although it is not crucial to the success of the method.\label{tabVqaCpAblation}}
\vspace{-8pt}
\end{table}

\begin{table*}[t]
\footnotesize
\renewcommand{\tabcolsep}{0.54em}
\renewcommand{\arraystretch}{1.25}
\centering
\begin{tabularx}{\linewidth}{X cccc c cccc}
\Xhline{1\arrayrulewidth}
                                          & \multicolumn{4}{c}{VQA-CP v2, Test set} & ~~~~~~~~ & \multicolumn{4}{c}{VQA v2, Validation set} \\ \cline{2-5} \cline{7-10}
                                          & Overall\hspace{2.1em} & Yes/no & Numbers & Other$\downarrow$ & ~ & Overall\hspace{2.1em} & Yes/no & Numbers & Other \\
\Xhline{1\arrayrulewidth}

SAN~\cite{yang2015stacked}  & 24.96\hspace{2.1em} & 38.35 & 11.14 & 21.74 & ~ & 52.02\hspace{2.1em} & -- & -- & -- \\
GVQA~\cite{agrawal2018don} & 31.30\hspace{2.1em} & 57.99 & 13.68 & 22.14 & ~ & 48.24\hspace{2.1em} & -- & -- & -- \\
Ramakrishnan \etal, 2018~\cite{ramakrishnan2018overcoming} & 42.04\hspace{2.1em} & 65.49 & 15.87 & 36.60 & ~ & 62.75\hspace{2.1em} & 79.84 & 42.35 & 55.16 \\
Grand and Belinkov, 2019~\cite{grand2019adversarial} & 42.33\hspace{2.1em} & 59.74 & 14.78 & 40.76 & ~ & 51.92\hspace{2.1em} & -- & -- & -- \\
RUBi~\cite{cadene2019rubi} & 47.11 \pp{0.51} & 68.65 & 20.28 & 43.18 & ~ & 61.16\hspace{2.1em} & -- & -- & -- \\
Teney \etal, 2019 \cite{teney2019actively} & 46.00\hspace{2.1em} & 58.24 & 29.49 & 44.33 & ~ & -- & -- & -- & -- \\
Product of experts~\cite{clark2019don} & 40.04\hspace{2.1em} & 43.39 & 12.32 & 45.89 & ~ & 63.21\hspace{2.1em} & 81.02 & 42.30 & 55.20\\ 
Clark \etal, 2019~\cite{clark2019don} & \textbf{52.01}\hspace{2.1em} & \textbf{72.58} & \textbf{31.12} & 46.97 & ~ & 56.35\hspace{2.1em} & 65.06 & 37.63 & 54.69\\ 
\hline
Our baseline model & 37.87 \pp{0.24} & 41.62 & 10.87 & 44.02 & ~ & 61.09 \pp{0.26} & 80.23 & 42.25 & 53.97\\
Proposed method & 42.39 \pp{1.32} & 47.72 & 14.43 & \textbf{47.24} & ~ & 61.08 \pp{0.12} & 78.32 & 42.16 & 52.81\\
\hline
Our baseline model ($\times$4 ensemble) & 39.30\hspace{2.1em} & 40.72 & 11.18 & 46.44 & ~ & 64.26\hspace{2.1em} & 82.07 & 44.56 & 56.33\\ 
Proposed method ($\times$4 ensemble) & 43.37\hspace{2.1em} & 47.82 & 14.35 & \textbf{49.18} & ~ & 63.47\hspace{2.1em} & 81.99 & 43.07 & 55.21\\ 

\Xhline{1\arrayrulewidth}
\end{tabularx}
\normalsize
\vspace{3pt}
\normalsize
\caption{Comparison with existing methods designed to improve generalization on VQA-CP (accuracy in percents). The evaluation on `yes/no' and `number' questions is highly unreliable (see \sect\ref{sec:vqaCp} and \cite{teney2020value}). On the `Other' questions however, our method surpasses all others. Our improvements on VQA-CP come only with a slight decrease in performance when trained and evaluated on the standard splits of VQA v2 (right columns). Reassuringly, the benefits of our method are cumulative with those of an ensemble (obtained by averaging the predictions of four models trained independently). The proposed method evaluated here uses environments built with question groups, $E$=15 environments, the relative variance regularizer, and no alternating optimization.\label{tabVqaCpComparison}
}
\vspace{-6pt}
\end{table*}

\subsection{Invariance to equivalent questions (GQA)}

\paragraph{Experimental setup.}
The GQA dataset~\cite{hudson2018gqa} is a VQA dataset built with images of the Visual Genome project~\cite{krishnavisualgenome} and questions generated from the scene graphs of these images.
The questions are generated from a large number of templates and hand-coded rules, such that they are of high linguistic quality and variety.
We present experiments that the annotations of ``equivalent questions'' that are provided with the dataset.
These annotations are not used in any existing model, to our knowledge.
A small fraction of training questions ($\sim$17.4\% in the balanced training set) are annotated with up to three alternative forms.
They involve a different word order or represent a different way of asking about a same thing.
For example:
\vspace{1pt}
\begin{itemize}
  \setlength{\parskip}{0pt}
  \setlength{\itemsep}{1pt}
\item[--] \textit{Is there a fence in the scene ?}\\
      \textit{Do you see a fence ?}\vspace{1pt}
\item[--] \textit{Which size is the green salad, small or large ?}\\
      \textit{Does the green salad look large or small ?}\vspace{1pt}
\item[--] \textit{Are there airplanes or cars ?}\\
      \textit{Are there any cars or airplanes in this photo ?}
\end{itemize}
\vspace{1pt}
Some alternative forms are already part of the dataset as other training questions, others are not.
The straightforward way to use these annotations is by data augmentation, \ie aggregating the equivalent forms with original training set.

\vspace{5pt}
\paragraph{Training environments with equivalent questions.}
We use our method to help learning invariance to the linguistic patterns of equivalent questions.
We use $E{=}4$ environments where we replace, in each, a question by its $e^\textrm{th}$ equivalent form if available, or the original form otherwise.
Each environment thus contains a single form of each question.

\begin{figure}[t]
  \centering
  \includegraphics[height=0.445\linewidth]{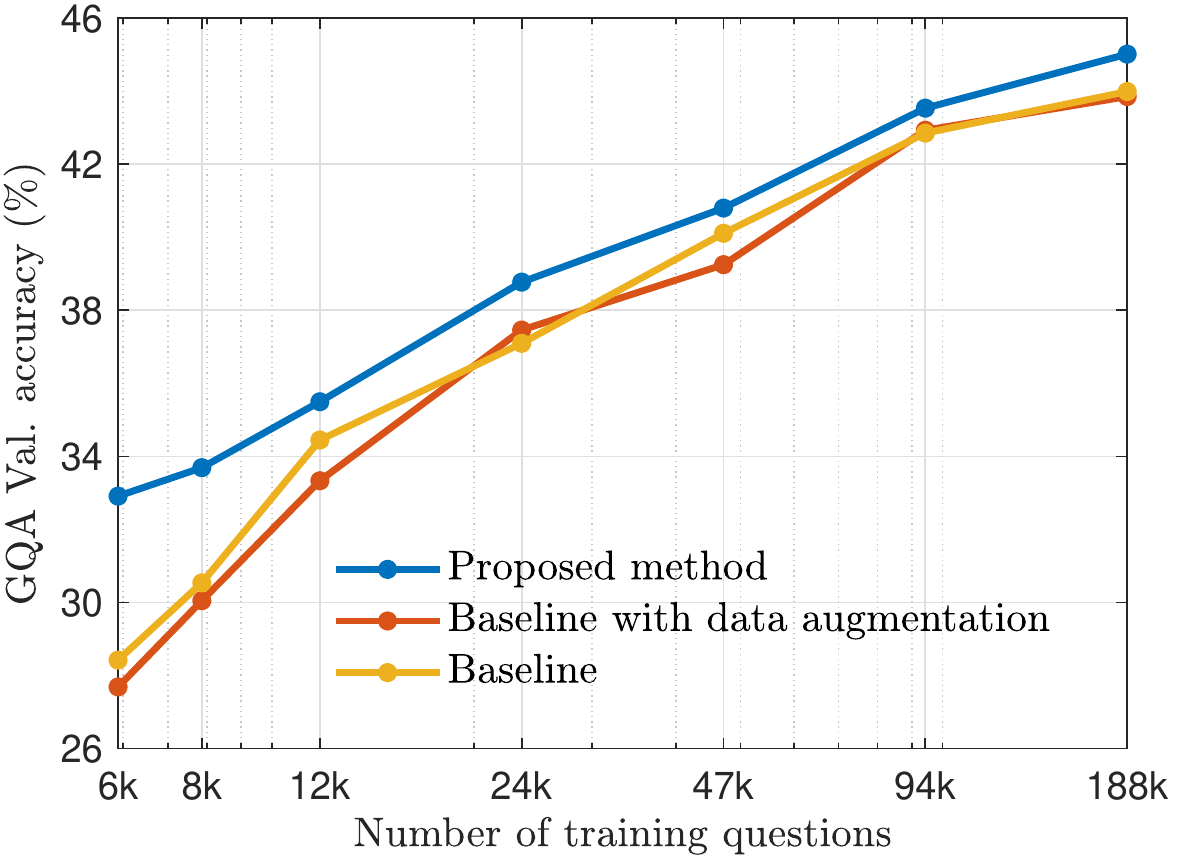}~\includegraphics[height=0.445\linewidth]{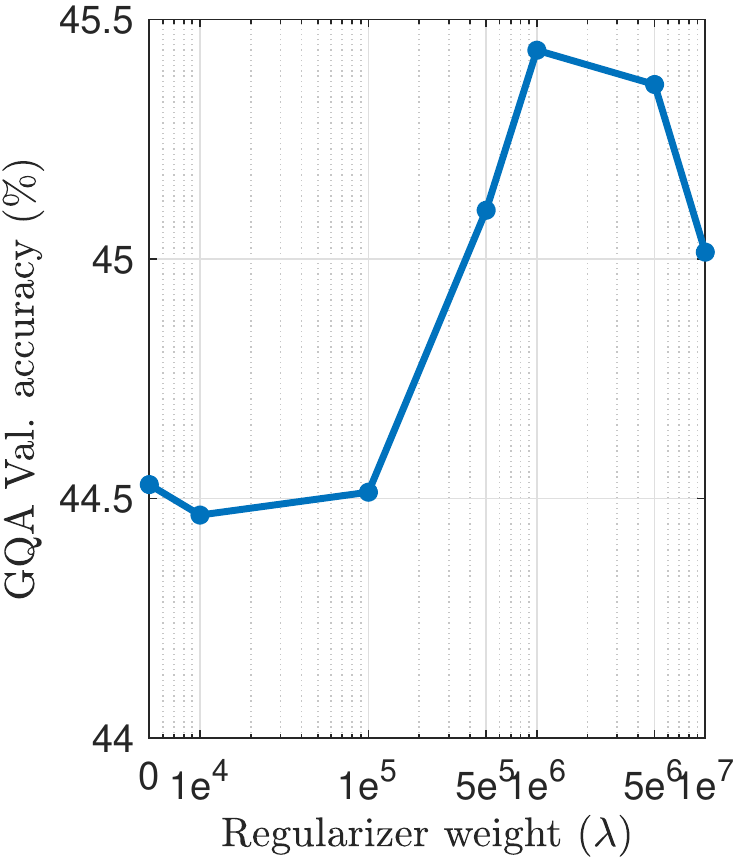}
  \caption{Experiments on GQA using equivalent questions to build environments. Our method provides consistent gains over the baseline, especially in the low-data regime. The improvement diminishes as more data is available, and is essentially imperceptible when the model is training on the full 2M training examples ($\sim$10$\times$ than shown on this plot). A naive use of the equivalent questions for data augmentation has a negative effect because it shifts the distribution of the training set away from the test set.\label{fig:gqa}}
  \vspace{-11pt}
\end{figure}

\vspace{5pt}
\paragraph{Results.}
We compare in \fig\ref{fig:gqa}a the accuracy of our method with same model trained on the standard training set, and with the data augmentation baseline described above. The data augmentation does not help despite the additional training examples, because it modifies the distribution of training examples away from the distribution of test questions. Our method, in comparison, brings a clear improvement. For a fair comparison, we made sure that the data augmentation uses the exact same questions (original and equivalent forms) in every mini-batch, such that the improvement is strictly brought on by the architectural differences of our method. The improvement with our method is greatest with low amounts of training data (we use random subsets of the full training set). The full dataset provides a massive 14M examples (about 1M in its balanced version), at which point the impact of our method is imperceptible. The training set then essentially covers the variety of linguistic forms and concepts exhaustively enough such that there is no benefit from the additional annotations.

It is worth noting that all improvements brought by our method come from only a small fraction of questions being annotated with equivalent forms. It would therefore be realistic to annotate a real VQA dataset with similar equivalent forms, and investigate possible gains with our method, which we hope to do in the future.

In \fig\ref{fig:gqa}b, we plot the accuracy as a function of the regularizer weight. The clear optimum confirms again that the regularizer is a crucial component of the method.

\subsection{Multi-dataset training (VQA v2~and~VG QA)}

\paragraph{Experimental setup.}
These experiments apply our method to the training of a model on multiple datasets simultaneously. The VQA v2 dataset has previously been aggregated with Visual Genome QA (VG)~\cite{krishnavisualgenome} as a simple way to use more training data. The datasets contain similar types of questions, but it is reasonable to assume that they have slightly different distributions. We use $E{=}2$ environments, the first one containing the VQA v2 training data, the second the VG data.

\vspace{5pt}
\paragraph{Results.}
In \tab\ref{tabVqaVg}, we compare our method with a model trained on VQA v2, another trained on VG, and one trained on the aggregation of the two datasets. The improvement is small but was verified over multiple training runs. We also ruled out explanation of the improvement as merely an ensembling effect, by comparing an ensemble of the baseline with one of the proposed method. The benefits of our method are cumulative with those of an ensemble, which suggests that our method should also apply to higher-capacity models. A number of such models have been described with a higher performance on VQA v2~\cite{chen2019uniter,gao2019dynamic,gao2019multi,li2019unicoder,liu2019learning,tan2019lxmert,yu2019deep} and it will be interesting to combine them with our method in the future.

\begin{table}[t]
\footnotesize
\scriptsize
\renewcommand{\tabcolsep}{0.10em}
\renewcommand{\arraystretch}{1.15}
\centering
\begin{tabularx}{\linewidth}{X cccccccc}
\Xhline{1\arrayrulewidth}
                                          & \multicolumn{6}{c}{VQA v2, Validation set} & ~ & VG\\ \cline{2-7} \cline{9-9}
                                          & Overall & ~ & Overall & Yes/no & Numbers & Other & ~ & Val.\\ \cline{2-2} \cline{4-9}
                                          & Ens.$\times$4 & ~ & \multicolumn{6}{c}{Single model}\\
\Xhline{1\arrayrulewidth}
Baseline model\\
~~~Trained on VQA v2			& 64.86 & ~ & 63.07 \pp{0.23}  & 81.40 & 42.09 & 54.21 & ~ & 49.67 \\
~~~Trained on VG				& 28.48 & ~ & 27.58 \pp{0.22}  & 0.11 & 36.03 & 47.11 & ~ & 60.17 \\
~~~Trained on Aggregated data		& 65.47 & ~ & 63.32 \pp{0.35}  & \textbf{82.27} & 40.99 & \textbf{55.98} & ~ & \textbf{61.20} \\
\hline
Proposed method\\
~~~Without variance reg.			& 64.33 & ~ & 62.18 \pp{0.27}  & 78.95 & 41.68 & 54.42 & ~ & 59.68 \\
~~~\textbf{With variance reg.}	& \textbf{65.73} & ~ & \textbf{63.80} \pp{0.17}  & 81.00 & \textbf{42.35} & 55.97 & ~ & 60.54 \\
\Xhline{1\arrayrulewidth}
\end{tabularx}
\normalsize
\vspace{3pt}
\normalsize
\caption{Multi-dataset training with VQA v2 and Visual Genome. The standard practice is to aggregate the two datasets. Our method treats them as two distinct training environments. The improvement is very small, but it comes at zero extra cost, and it was verified over multiple runs (mean and standard deviation are reported), as well as in an ensemble (first column). It was also verified on two different implementations of the baseline model (not in table).\label{tabVqaVg}}
\vspace{-5pt}
\end{table}

\section{Conclusions}

We presented a method to train a deep models to better capture the mechanism of a task of interest, rather than blindly absorbing all statistical patterns from a training set. The method is based on the identification of correlations that are stable across multiple training environments, \ie subsets of the training data.
We described several strategies to build these environments using different forms of prior knowledge and auxiliary annotations. We showed benefits in various conditions including out-of-distribution test data, low-data training, and multi-dataset training.

An exciting challenge in computer vision is to design models solving tasks rather than datasets. Our strong results on VQA, which is known for its challenges in generalization and data scarcity, give us confidence that suitable tools like this method are emerging to make progress in this direction.

The proposed method is related to a series of works on IRM that appeared concurrently with the preparation of this paper~\cite{arjovsky2019invariant,ahuja2020invariant,chang2020invariant,Choe2020AnES,idnanilearning}. Much remains to be done in terms of theoretical and empirical comparisons of these methods and their application to real (non-toy) data beyond VQA.

\clearpage

{\small\bibliographystyle{ieee}\bibliography{Bibliography}}
\clearpage

\appendix
\onecolumn
\section*{Supplementary material}
\vspace{12pt}

\section{Implementation of the VQA model}
\label{appendix:vqa}

The VQA model used in our experiment follows the general description of Teney \etal~\cite{teney2017challenge}. We use the ``bottom-up attention'' features~\cite{anderson2017features} of size 36$\times$2048, pre-extracted and provided by Anderson \etal\footnote{~https://github.com/peteanderson80/bottom-up-attention} The non-linear operations in the network use gated hyperbolic tangent units. The word embeddings are initialized as GloVe vectors~\cite{pennington2014glove} of dimension 300, then optimized with the same learning rate as other weights of the network. All activations except the word embeddings and their average are of dimension 512. The answer candidates are those appearing at least 20 times in the VQA v2 training set, \ie a set of about 2000 answers. The output of the network is passed through a logistic function to produce scores in $[0,1]$. The final classifier is trained from a random initialization. The model is trained with backpropagating a binary cross-entropy loss, and updating all weights with AdaDelta.

We use early stopping in all experiments to prevent overfitting. When using a distinct validation and test set, we report the accuracy on the test set, at the epoch of highest accuracy on the validation set.

\section{Implementation of the proposed method}

In our experiments with VQA-CP, the environments are built using either the ground truth question types, or an unsupervised clustering of the training questions. In the latter case, we use the $k$-means algorithm on a bag-of-words representations of questions. These representations are binary vectors whose length is equal to the size of the vocabulary of words that appear $\geq$10 times in the training set. Each component of the vector is equal to one if the corresponding word is present in the question, or zero otherwise. The clustering algorithm uses the cosine similarity as a metric. We also experimented with clustering representation of the questions made of their average GloVE embeddings~\cite{pennington2014glove} but the results were slightly worse.

The alternating optimization scheme showed a slight improvement in accuracy on VQA-CP. However, it brings another tunable hyperparameter (the number of warm-up epochs). We did not use it in most experiments because of the low potential return compared to the added expense in compute for tuning this hyperparameter. We have not verified whether the improvement holds on datasets other than VQA-CP.

\section{Additional experiments and negative results}
This section provides some insights on the timeline of the experiments presented in the paper, and of others that brought negative results.

Our initial, most encouraging results were obtained with VQA-CP, using the ground truth annotations of question types. The question types are known to be spuriously correlated with the answers across the training and test sets of VQA-CP, by construction of the dataset~\cite{agrawal2018don}. The use of this very fact is specific to the VQA-CP dataset, and it somewhat defeats the very objective of VQA-CP of encouraging generalizable models. Other recent works have used these annotations however~\cite{clark2019don}, so it seemed fair game to do so as well. Nonetheless, we wanted to demonstrate a more general usage of our method that did not rely on these annotations. We experimented with various strategies to build environments by clustering the training data. The one presented in the paper simply uses the questions, which essentially approximates the labeling of the question groups. We tested other strategies, all of which proved unsuccessful, both on in- and out-of-distribution test data. We tried to cluster the training data based on the answers, the question words, the image features, and all combinations thereof.

With the GQA dataset, we experimented with using two environments, where we would sample, in the first, from the standard balanced training set, and in the second, from the larger unbalanced training set. The accuracy did however decrease on the standard balanced validation set.

The experiments we considered for this paper focused on VQA, but we believe there are a lot of possible other applications worth exploring, well beyond tasks in vision-and-language.

At test time, we use the average of the classifier weights learned across the training environments. We tried other strategies, such as using the median values, but the difference was insignificant. The variance regularizer already brings the weights to very similar values across environments.

\begin{table*}[hb!]
  \caption{Accuracy per question type on the GQA dataset~\cite{hudson2018gqa}. Almost all categories benefit equally from the proposed method.\label{tabComparative}}
  \vspace{6pt}
  \centering
  \renewcommand{\tablefont}{\footnotesize}
  \tablefont
  \renewcommand{\arraystretch}{1.00}
  \begin{tabular}{lccccccccccc}
    \toprule
    ~ & Overall & Verify & Query & Choose & Logical & Compare & Object & Attribute & Category & Rel. & Global\\
    \midrule
    \multicolumn{12}{l}{With 6k Training examples (leftmost points on \fig\ref{fig:gqa}a)}\\
    Baseline & 28.42 & 50.00 & 14.61 & 22.67 & \textbf{51.58} & 45.67 & \textbf{52.31} & 32.84 & 17.93 & 23.02 & 23.57 \\
    Baseline with data augmentation& 27.69 & 51.07 & 13.48 & 23.91 & 49.03 & 44.31 & 47.17 & 33.03 & 15.32 & 22.60 & 17.20 \\
    \textbf{Proposed method} & \textbf{32.91} & \textbf{52.26} & \textbf{20.89} & \textbf{29.50} & {50.42} & \textbf{50.76} & \textbf{52.31} & \textbf{37.64} & \textbf{25.85} & \textbf{26.91} & \textbf{35.03} \\
    \midrule
    \multicolumn{12}{l}{With 188k Training examples (rightmost points on \fig\ref{fig:gqa}a)}\\
    Baseline & 43.99 & 60.48 & 32.73 & 52.97 & 57.68 & \textbf{51.95} & 67.87 & 47.05 & 37.86 & 38.56 & \textbf{52.87} \\
    Baseline with data augmentation & 43.85 & 60.17 & 32.74 & 52.52 & \textbf{58.18} & 49.41 & \textbf{69.67} & 46.97 & 37.68 & 38.18 & 49.68 \\
    \textbf{Proposed method} & \textbf{45.01} & \textbf{61.55} & \textbf{34.15} & \textbf{55.18} & {57.74} & {48.90} & {68.64} & \textbf{48.28} & \textbf{41.60} & \textbf{38.97} & {49.04} \\
    \bottomrule
  \end{tabular}
  \vspace{-8pt}
\end{table*}

\end{document}